\documentclass{edm_article}
\usepackage[T1]{fontenc}
\usepackage{graphicx}
\usepackage{amssymb}
\usepackage{amsmath}
\usepackage{tikz}
\usepackage{subcaption}
\usepackage{url}
\usepackage{booktabs}
\usepackage{placeins}
\usetikzlibrary{positioning,arrows.meta,calc}

\begin{document}
%
% \title{UNVaMP: A Modular Variational Deep Learning Architecture for Knowledge Tracing}
\title{UNVaMP: Neural Knowledge Tracing with Variational Regularization of Latent Knowledge Dynamics}
%
% \author{
% \IEEEauthorblockN{Carson J. Cook, Ahmed J. Zerouali, Anthony Schmidt, Reginald Ziedzor, Paul Lin, and Luke G. Eglington}
% % \IEEEauthorblockA{
% % \\ Amplify Education, Brooklyn, NY, USA \\
% % Email: \{ccook, aschmidt, rziedzor, leglington\}@amplify.com
% % }
% }

\numberofauthors{6}
\author{
%
%
% 1st. author
\alignauthor Carson J. Cook\\
       \affaddr{Amplify Education, Inc.}
       \email{ccook@amplify.com}
% 2nd. author
\alignauthor Ahmed J. Zerouali\\
       \affaddr{Amplify Education, Inc.}
       \email{azerouali@amplify.com}
% 3rd. author
\alignauthor Anthony Schmidt\\
       \affaddr{Amplify Education, Inc.}
       \email{aschmidt@amplify.com}
\and  % use '\and' if you need 'another row' of author names
% 4th. author
\alignauthor Reginald Ziedzor\\
       \affaddr{Amplify Education, Inc.}
       \email{rziedzor@amplify.com}
% 5th. author
\alignauthor Paul Lin\\
       \affaddr{Amplify Education, Inc.}
       \email{plin@amplify.com}
% 6th. author
\alignauthor Luke G. Eglington\\
       \affaddr{Amplify Education, Inc.}
       \email{leglington@amplify.com}
}

% \numberofauthors{1}
% \author{
% %
% %
% % 1st. author
% \alignauthor Anonymous\\
%        \affaddr{Anonymous Institution}\\
%        \affaddr{anonymous@anonymous.edu}
% }

\maketitle

\begin{abstract}
% Present: what the paper does
% Past: what we found

We introduce the Unified Neural Variational Measurement of Proficiency (UNVaMP) architecture, a knowledge tracing method that integrates observed student-item interactions with internal memory to produce evolving latent representations of student knowledge. These representations support accurate predictions of future responses while enabling explicit control over the smoothness of estimated learning trajectories. UNVaMP can be configured as either a purely neural model or a hybrid model that predicts responses through an interpretable measurement function over the latent space. We show that a pure neural configuration (UNVaMP-MLP) achieves the strongest predictive performance among compared models on three out of four datasets. Meanwhile, a hybrid configuration (UNVaMP-MIRT, using a 1PL MIRT measurement function) lags only slightly behind UNVaMP-MLP, indicating that the predictive cost of interpretability is modest.

Beyond predictive accuracy, UNVaMP provides the following: a principled mechanism for controlling volatility when estimating student latent variables, quantification of uncertainty over student knowledge state estimates, and flexible input specification that supports heterogeneous student-item interaction features. In addition, the hybrid UNVaMP-MIRT configuration generates interpretable moment-in-time student knowledge state estimates. Using an experimental dataset, we show that auxiliary inputs induce structured changes in the predictive behavior of UNVaMP-MIRT, consistent with sensitivity to underlying structure beyond response correctness. Furthermore, through a simulation study, we show that UNVaMP yields well-behaved knowledge state estimates under controlled measurement conditions. In total, these results indicate that UNVaMP is both practically useful for real-world education systems and capable of recovering meaningful underlying structure from student-item interactions.

\end{abstract}


\section{Introduction}
% Present: background and framing
% Past: specific prior work

Building models of student learning often involves balancing competing needs for rigorous student knowledge estimation and flexible model fitting over large, sparse, and noisy datasets, while recognizing that student knowledge evolves with each interaction. In contrast, classical measurement models are typically designed for tightly controlled test environments (e.g., standardized tests) and impose strict assumptions about student latent knowledge structures. Knowledge tracing (KT) models seek to incorporate sequential (or temporal) information about student-item interactions and fall into two broad categories: those rooted in classical statistical frameworks~\cite{corbett-1995-bkt,lkt, pelanek-2016-elo}, and those built upon modern deep learning methods~\cite{choi-2020-saint, pandey-2019-sakt, piech-dkt-2015}. Classical approaches often preserve much of the framing of traditional measurement models but relax certain constraints to capture learning dynamics and may support less controlled settings. Modern deep learning approaches to KT typically relax both statistical and latent knowledge assumptions in pursuit of flexibility and predictive power, although hybrid and theory-informed methods are increasingly prevalent in the literature.

In the remainder of this section, we briefly summarize the measurement and KT paradigms, followed by a discussion of existing KT methods, along with their strengths and limitations, then distill from this analysis a set of architectural desiderata outlining desirable properties for modern KT architecture designs. The UNVaMP architecture is then described in detail in Section 2. We then describe our datasets and experiments in Section 3, followed by a presentation of our results in Section 4. Finally, we summarize our findings and discuss limitations and future work in Section 5.

% \paragraph{Notation}
% All latent variables and internal representations (e.g., embeddings, extracted features) are vector-valued unless otherwise specified. Scalars are explicitly identified where needed.

\subsection{Classical Measurement Models}
In classical measurement, models are built around an underlying construct of interest, typically represented by one or more latent variables that are assumed to be static within the measurement window~\cite{borsboom2008latent}. In such models, student and item parameters are estimated directly, often via maximum likelihood, the expectation-maximization (EM) algorithm, or related variational approaches, rather than through amortized inference \cite{hambleton1991fundamentals_irt}. Although there are limitations with these models in terms of predictive performance and flexibility, they are popular due to their interpretability, well-understood statistical properties, and mature implementations. Two very popular classical frameworks are Item Response Theory (IRT) and Cognitive Diagnostic Modeling (CDM). We briefly discuss IRT below, but point readers towards other materials discussing the properties and usage of CDMs~\cite{rupp-2010-diagnostics, templin-2013-cdm}. 

\subsubsection{Item Response Theory}
IRT is a classical measurement framework in which observed student-item interactions are modeled as a probabilistic function of latent student proficiency and item characteristics~\cite{hambleton1991fundamentals_irt}. Multidimensional IRT (MIRT) extends this framework by defining latent proficiency as a vector, where each dimension represents a latent skill~\cite{reckase200618_mirt}. For instance, the two-parameter logistic (2PL) MIRT model is formulated as
\[
    P\left[ Y_{i,j}=1 \mid \theta_i, a_j, d_j \right] = \frac{1}{1 + \exp\{-(a_j^\top \theta_i + d_j)\}} ,
\]
where $Y_{i,j} \in \{0,1\}$ denotes an incorrect or correct student response to item $j$, $\theta_i$ is the latent proficiency vector for student $i$, $a_j$ is the vector of item \textit{discrimination} (or \textit{loading}) parameters, and $d_j$ is a scalar item intercept related to that item's \textit{difficulty}. The vectors $\theta_i$ and $a_j$ have one dimension per latent skill. The one-parameter logistic (1PL) MIRT model, also known as the multidimensional Rasch model, is a restricted case of 2PL MIRT where the dimensions of $a_j$ have fixed values (typically 0 or 1).

\subsection{Knowledge Tracing}
Knowledge tracing is a broad category of methods that, unlike classical measurement models, treat student knowledge as evolving throughout a sequence of student-item interactions~\cite{shen-2021-knowledgetracing}. We group KT approaches into two broad categories: those grounded in classical measurement philosophies or statistical frameworks, and those built on modern deep learning methods. In our discussion of deep learning knowledge tracing (DLKT), we examine approaches that focus primarily on prediction, as well as hybrid methods that incorporate elements of classical measurement.

\subsubsection{Classical and Statistical KT Methods}
Classical and statistical KT methods typically model student knowledge directly and may utilize a longitudinal or dynamic adaptation of a classical measurement model. Popular families include Bayesian Knowledge Tracing (BKT)~\cite{corbett-1995-bkt,baker-2008-bkt}, Elo~\cite{gonzalez-2014-elo,pelanek-2016-elo}, and various dynamic IRT and CDM models~\cite{verhelst-1993-dynamicRasch,wang-2013-dynamicIRT,wang-2019-longIRT,lee-2025-dynamicIRT,kaya-2016-dynamicCDM,chen-2018-dynamicCDM,wang-2018-dynamicCDM}. Many variants exist, but all primarily consist of estimating student and item parameters given a sequence of student-item interactions.

Within classical KT, another well-studied and typically high-performing family of models is Logistic Knowledge Tracing (LKT)~\cite{pavlik2021LKTpaper}. LKT is a framework that attempts to unify the many previous KT models that use logistic regression as their core predictive model. These prior models typically use hand-engineered features that could in principle be used together but were not (e.g., AFM \cite{cen2006AFM1}, PFA \cite{pavlik2009PFA}, variants of PFA~\cite{gong2010comparing}, R-PFA \cite{galyardt2015RPFA}, spacing models like PPE \cite{walsh2018PPE1,walsh2018PPE2}, and other features inspired by cognitive psychology \cite{pavlik2008optimalspacing}). However, selecting among these candidate features is challenging, so feature selection using lasso regression was included in the framework. The lasso regression approach was recently found to outperform several popular deep learning approaches on four benchmark datasets \cite{pavlik2023lassolkt}.

Despite their conceptual clarity, classical methods often suffer from several major shortcomings. First, longitudinal variants of IRT and CDM often require re-fitting student (and possibly item) parameters to obtain updated estimates given new observations \cite{wang-2019-longIRT}. Other variants of these IRT and CDM methods, along with models like Elo \cite{pelanek-2016-elo}, support using updating rules for real-time parameter re-estimation, but these require analytical derivations or heuristics that limit model expressiveness, and these methods may still struggle to scale with larger datasets \cite{lindsey2014colt}. In addition, when many skills are present (e.g., dozens or hundreds), especially in large datasets, CDM and IRT methods may fail to converge without imposing significant assumptions or structural simplifications. Finally, classical approaches typically do not flexibly support auxiliary inputs beyond minimal predictors (e.g., sequential binary responses). Those that do, such as explanatory IRT or LKT models, utilize manually engineered features and require strong assumptions about functional relationships between inputs and outputs~\cite{Wilson2004}. This manual approach can become cumbersome as new data sources become readily available (e.g., open-ended text responses), requiring significant resources to develop new features or adapt existing ones. This limitation extends to features intended to act as latent variables representing student knowledge, such as those intended to behave similarly to Elo parameters~\cite{pavlik2025evolutionary}, since their interpretation will be confounded by the inclusion of new auxiliary inputs.

% This manual approach can become cumbersome and time-consuming as new data sources become readily available (e.g., open-ended responses), requiring significant resources to develop new features or adapt existing ones for use by the model. This limitation extends to features intended to act as latent variables to represent student knowledge. Although some features behave similarly to the Elo parameters that are sometimes used to represent student knowledge \cite{pavlik2025evolutionary}, the interpretation of these features would be confounded by the inclusion of new auxiliary inputs and could require a redesign of the feature.

\subsubsection{DLKT Methods}
% In response to the limitations of classical approaches, a large number of modern DLKT methods have been introduced in the past decade. Some architectures focus primarily on obtaining the best predictive accuracy possible, while others take a hybrid approach that seeks to model one or more of student knowledge, item properties, learning parameters, in an interpretable way.

In response to the limitations of classical methods, numerous DLKT architectures have emerged over the past decade, ranging from those focused primarily on obtaining maximum predictive accuracy, to hybrid approaches that incorporate interpretable representations of student knowledge, item properties, or learning dynamics.

% Big list:
% nakagawa-2019-gkt, zhang-2016-dkvmn, ghosh-2020-akt, pandey-2019-sakt, yang-2020-gikt, choi-2020-saint, ruan-2021-vdkt, cheng-2025-uncertaintyAwareKT, christie-2024-lens, yeung-2018-dktPlus, guo-2021-atkt, shin-2020-saintPlus
Prediction-focused DLKT architectures generally forgo interpretable student or item representations, limiting their scope to next-response prediction in sequences of student-item interactions. The first major example was DKT \cite{piech-dkt-2015}, a recurrent architecture focused exclusively on prediction. Following its introduction, numerous DLKT models have been proposed, many achieving state-of-the-art performance at the time of publication. Prominent examples include SAKT~\cite{pandey-2019-sakt}, AKT~\cite{ghosh-2020-akt}, and SAINT~\cite{choi-2020-saint}. Within the prediction-focused DLKT paradigm, various extensions have been explored, such as incorporating uncertainty estimation~\cite{cheng-2025-uncertaintyAwareKT, christie-2024-lens, ruan-2021-vdkt}. DLKT models can exhibit volatility and unintuitive behavior in student learning dynamics~\cite{khajah-2016-howdeepiskt,yeung-2018-dktPlus,wang-2022-wrongDKT,zhang-2022-modifying,yin-2023-dtransformer}. Proposed remedies have included penalization of output-space inconsistency~\cite{yeung-2018-dktPlus}, attention mechanisms~\cite{wang-2022-wrongDKT}, and contrastive training procedures~\cite{yin-2023-dtransformer}. Despite these advances, such models remain largely uninterpretable and focused on prediction.

% Big list:
% yeung-2019-deepIRT, pandey-2020-rkt, wang-2021-hawkeskt, chen-2023-qikt, wang-2023-dcd, zhou-2025-dkt2
To address these interpretability limitations, several hybrid architectures incorporating elements of classical measurement, such as IRT- or CDM-based outputs, have been proposed~\cite{wang-2019-neuralcd, wang-2020-varfa, wu-2020-vibo, song-2023-cmncd}. Many early hybrid methods assumed static data generation and did not model learning dynamics. More recently, dynamic hybrid architectures suitable for KT have emerged, incorporating interpretability constraints such as IRT-style response functions~\cite{yeung-2019-deepIRT, chen-2023-qikt, zhou-2025-dkt2}, CDM-inspired interaction functions~\cite{wang-2023-dcd}, or structured dynamics~\cite{pandey-2020-rkt, wang-2021-hawkeskt}. However, to our knowledge, methods that introduce interpretability into the prediction mechanism do not tend to incorporate accompanying dynamics regularization~\cite{yeung-2019-deepIRT,chen-2023-qikt}, while methods that impose structure on dynamics do not produce interpretable knowledge states~\cite{pandey-2020-rkt,wang-2021-hawkeskt}, and uncertainty representation is often left unaddressed.

\subsection{Architecture Desiderata}
\label{sec:desiderata}
Although knowledge tracing models are often judged primarily by predictive accuracy, prior work has argued that additional attributes are critical for practical usability~\cite{rachatasumrit2024interpretability}. We propose a set of architectural desiderata to guide DLKT architecture design and address common challenges with existing approaches. While many models satisfy subsets of these desiderata, we are unaware of any that explicitly address each of them.

\subsubsection{Regularization of Dynamics}
The volatility issues that have been observed in the DLKT paradigm may hamper their usability in many real-world contexts. In deployed systems, models with poorly constrained dynamics can yield unstable proficiency estimates that undermine reliability and interpretability. Accordingly, DLKT methods should incorporate explicit regularization mechanisms that promote smooth, stable trajectories of the inferred student knowledge states.

\subsubsection{Interpretable Knowledge States}
Although many DLKT architectures exist, few output interpretable knowledge states. Full transparency into all internal mechanisms (e.g., dynamics and response prediction) is ideal but impractical in DLKT without imposing constraints that undermine the benefits of deep learning. Instead, we suggest that moment-in-time summaries of student knowledge are sufficient for many practical use cases, provided the underlying dynamics are well-regularized, even if those dynamics are not themselves interpretable. To date, many hybrid methods either provide only partial knowledge state interpretability, or are too constrained to fully leverage DLKT's expressiveness. Thus, a model that allows users to specify an interpretable latent knowledge representation while retaining flexibility in modeling student learning dynamics would be highly useful.

\subsubsection{Uncertainty Quantification}
A common challenge for DLKT models is the absence of distributional and structural assumptions that underpin classical uncertainty quantification, making it difficult to define appropriate standards for uncertainty updating and calibration in practice. Nevertheless, a DLKT model should provide an internally consistent representation of uncertainty that reflects changes in its confidence as evidence accumulates. Even without statistical guarantees, such uncertainty estimates have practical utility, such as revealing diminishing returns as data accrue, informing when inference outputs should be surfaced, and enabling diagnostic analyses related to external signals such as proportion correct or recency.

\subsubsection{Flexibility of Inputs}
A major benefit of deep learning is that it relaxes two onerous requirements of classical modeling: pre-specifying functional relationships between inputs and outputs, and hand-engineering features to increase model capacity. Accordingly, a DLKT model should be capable of incorporating arbitrary sets of inputs, provided they do not introduce information leakage with respect to the response. While classical methods can, in principle, be expanded to accommodate additional inputs (e.g., via additional coefficients), this flexibility is largely nominal in practice. Such approaches scale poorly to large feature spaces, and still require pre-specified functional forms (e.g., linear effects or interaction terms).

\section{The UNVaMP Architecture}

\begin{figure*}[!tbp]
    \centering
    \includegraphics[width=6.75in]{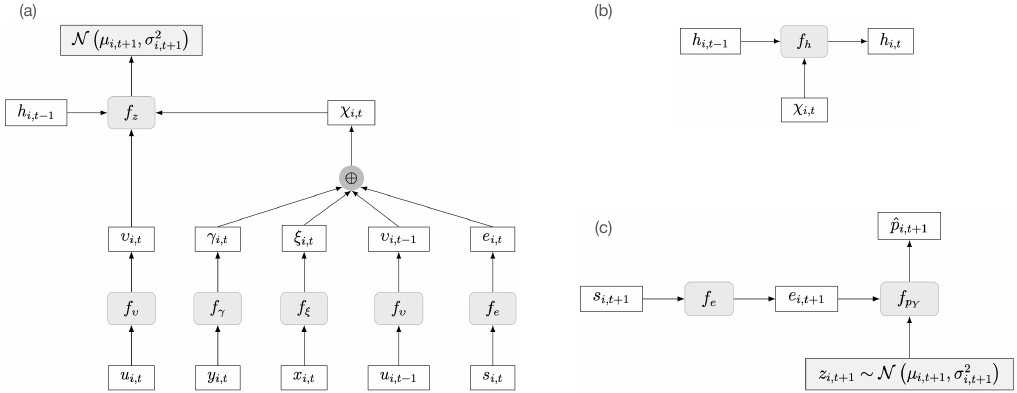}
    \caption{
        Overview of the UNVaMP architecture.
        Panel \textbf{(a)} shows the \textit{inference model}, which combines internal memory about student $i$ with extracted features from their most recent interaction to obtain an estimated latent knowledge distribution.
        Panel \textbf{(b)} shows the \textit{memory model}, which combines the previous internal memory about student $i$ with extracted features from their most recent interaction to obtain an updated internal memory.
        Panel \textbf{(c)} shows the \textit{response model}, which receives latent variable samples from the knowledge state distribution for student $i$ and generates predicted responses to items.
    }
    \label{fig:unvamp-architecture}
    \Description{
        This figure shows an overview of the UNVaMP architecture, which is composed of three components: an inference model, a memory model, and a response model. The diagram illustrates how student interaction data are processed over time to produce latent knowledge state estimates and predicted responses.
        
        Panel (a) depicts the inference model. Encoded features from the student's most recent interaction, including response information, item metadata, and other features, are combined with the student's prior internal memory state. These inputs are used to generate a latent knowledge state distribution for the next time step, represented as a Gaussian distribution with a mean and variance.
        
        Panel (b) shows the memory model. The previous hidden memory state and the current interaction features are combined to update the internal memory representation, which preserves information about the student’s past interactions and is used for subsequent inference.
        
        Panel (c) shows the response model. A sample (or point estimate, during inference) derived from the latent knowledge state distribution is combined with item embeddings (parameters) to generate a predicted probability of correctness for the next item.
    }
\end{figure*}

We introduce the Unified Neural Variational Measurement of Proficiency (UNVaMP) architecture. Inspired by dynamic deep learning architectures that encode sequential inputs as latent distributions~\cite{christie-2024-lens,girin-2021-dynamicVAEReview}, UNVaMP combines a memory model and an inference model (encoder) to produce moment-in-time latent knowledge state distributions, which are passed through a response model to predict student-item interaction outcomes. The response model thus serves as a modular decoder that determines how latent distributions map to predictions, and whether those latent distributions carry interpretable meaning. When configured with a neural decoder (UNVaMP-MLP), the latent space has no prescribed interpretation; when configured with a 1PL MIRT measurement function (UNVaMP-MIRT), the latent dimensions are directly interpretable as per-skill proficiency estimates, and item embeddings correspond to skill loadings and intercepts.

In UNVaMP, latent dynamics are regularized via a Kullback-Leibler divergence penalty between temporally adjacent distributions. When the response model is an interpretable measurement function, UNVaMP yields knowledge state estimates with prescribed meaning. Because inputs and memory are encoded as latent distributions, uncertainty representations are available through distributional parameters. All configurations support arbitrary auxiliary inputs. Use of the term ``variational'' refers to regularization of latent dynamics via a distributional divergence penalty, not variational Bayesian inference. Fig.~\ref{fig:unvamp-architecture} provides a full diagram of the architecture.

\subsection{Inference Model and Memory Model}
For a given student, the \textit{inference model} (or encoder) $f_z$ uses extracted features and hidden states---representing the model's internal memory---to output the mean and variance of a Gaussian distribution over the student's latent proficiency at a given time point. The hidden states are the outputs of the \textit{memory model} $f_h$, a recurrent neural network (RNN) that uses extracted features to build an internal hidden memory representation of the student; see Fig.~\ref{fig:unvamp-architecture}(a)--(b).

For student $i$ with a sequence of student-item interactions at time points $t = 1,2, ..., T_i$, the following inputs are available for each $t$:
\[
    \left( y_{i,t}, s_{i,t}, x_{i,t}, u_{i,t} \right),
\]
where $y_{i,t}$ denotes the observed response data (e.g., correct/incorrect) and $s_{i,t}$ is an identifier (typically an integer index) for the item the student interacted with. The variable $x_{i,t}$ reflects any auxiliary data pertaining to the student-item interaction (e.g., item or skill metadata), and $u_{i,t}$ represents pre-response data that are measured before the next response $y_{i,t+1}$ is observed, such as the time-gap between $t$ and $t+1$. The $x_{i,t}$ and $u_{i,t}$ inputs are optional, and can be flexibly specified, so long as they do not leak information about the response at $t+1$. In this study, predictions target only the binary correct/incorrect score at $t+1$.

During the forward pass, the inputs are sent through \textit{feature extractors} $f_{\gamma}$, $f_{\xi}$, $f_{\upsilon}$, and an \textit{item embedder} $f_{e}$ to obtain:
\[
\begin{alignedat}{2}
    \gamma_{i,t} &= f_{\gamma}(y_{i,t}), \quad
    & \xi_{i,t} &= f_{\xi}(x_{i,t}) \\
    \upsilon_{i,t} &= f_{\upsilon}(u_{i,t}), \quad
    & e_{i,t} &= f_{e}(s_{i,t}) \;,
\end{alignedat}
\]
Where $\gamma_{i,t},\ \xi_{i,t},\ \upsilon_{i,t},\ e_{i,t} \in \mathbb{R}^d$ and $d$ is a specified (or tuned) embedding/encoding dimensionality. The representations $\gamma_{i,t}$, $\xi_{i,t}$, and $e_{i,t}$, along with $\upsilon_{i,t-1}$ (the extracted pre-response features from $t-1$), are then concatenated to obtain:
\[
    \chi_{i,t} = [ \gamma_{i,t}, \xi_{i,t}, \upsilon_{i,t-1}, e_{i,t} ] .
\]
The representation $\chi_{i,t}$ describes the full set of data available immediately after observing $y_{i,t}$. The hidden state $h_{i,t-1}$ is obtained from the memory model $f_h$ using $\chi_{i,t-1}$ and $h_{i,t-2}$:
\[
    h_{i, t-1} = f_{h}(\chi_{i,t-1}, h_{i,t-2}) ,
\]
where $h_{i,t-1}$ reflects UNVaMP's internal memory about student $i$, prior to observation $t$.

Then, $\chi_{i,t}$, $\upsilon_{i,t}$, and $h_{i,t-1}$ are passed to the inference model $f_{z}$, which outputs a Gaussian distribution over student latent variable $z_{i,t+1}$:
\[
    \mathcal{N}(\mu_{i,t+1},\sigma_{i,t+1}^2) = f_\mathrm{z}\left(\chi_{i,t}, \upsilon_{i,t}, h_{i,t-1} \right) ,
\]
where $\mu_{i,t+1}$ represents the expected value of $z_{i,t+1}$, and $\sigma_{i,t+1}^2$, a vector representing the diagonal of a covariance matrix, reflects the model's uncertainty about $z_{i,t+1}$. Here, we have that $ \mu_{i,t+1},\ \sigma_{i,t+1}^2,\ z_{i,t+1} \in \mathbb{R}^k$, where $k$ is the latent dimensionality. The $t+1$ subscripts denote that these values represent the model's belief about what the student's knowledge state will be at the time of response $t+1$. During training, $z_{i,t+1}$ is sampled from $\mathcal{N}(\mu_{i,t+1},\sigma_{i,t+1}^2)$ in the forward pass, while during inference, it is given by point estimate $z_{i,t+1}=\mu_{i,t+1}$. 

We conclude our discussion of the inference model $f_z$ and memory model $f_h$ by noting several design choices. First, features $\chi_{i,t}$ are passed directly to $f_{z}$, rather than via $f_{h}$. This restricts the role of $f_{h}$ to that of a memory mechanism only, preventing it from being the conduit of new information at time point $t$. Second, the latent variable $z_{i,t+1}$ is never ingested by $f_h$, imposing a separation between the internal representations learned by $f_h$ and the point-in-time summaries given by $f_z$. In doing so, the internal expressiveness of the memory and inference models is left unconstrained by the latent distributions they produce. This separation has the added computational benefit of allowing the memory model to operate independently of the inference model's outputs during the forward pass, permitting the use of optimized sequence processing without the bottleneck of sequentially ingesting sampled latent variables.

\subsection{Response Model}
The \textit{response model} $f_{Y}$, shown in Fig.~\ref{fig:unvamp-architecture}(c), maps student latent knowledge states, along with item embeddings, to predicted response probabilities.

Once the student latent variable $z_{i,t+1}$ is obtained from $f_z$ and the next-item embedding $e_{i,t+1}$ is obtained from the embedder $f_e$ via $e_{i,t+1} = f_e\left( s_{i,t+1} \right)$, these values are passed into $f_{Y}$ to obtain:
\[
    \hat{p}_{i,t+1} = f_Y \left( z_{i,t+1}, e_{i,t+1} \right) ,
\]
where $\hat{p}_{i,t+1}$ is the scalar predicted probability that student $i$ will respond correctly to item $s_{i,t+1}$ at time point $t+1$, conditioned on the latent variable $z_{i,t+1}$ and the item embedding $e_{i,t+1}$. The items $s_{i,t+1}$ are known during training, and can be chosen arbitrarily during inference.

When $f_Y$ is a known measurement function, it determines the interpretation of the latent distribution. For the 1PL MIRT configuration used in this study, the item embedder $f_e$ outputs interpretable embeddings
\[
    e_{i,t+1}=\left( a_{s_{i,t+1}}, d_{s_{i,t+1}} \right) ,
\]
so that:
\[
    \hat{p}_{i,t+1} = \frac{1}{1 + \exp\{-(a_{s_{i,t+1}}^\top z_{i,t+1} + d_{s_{i,t+1}})\}} .
\]
Here, the ``embedding'' is a tuple containing $a_{s_{i,t+1}}$ and $d_{s_{i,t+1}}$, which represent the loadings and intercept for item $s_{i,t+1}$, where $d_{s_{i,t+1}}$ can either be learned during fitting or supplied as data. In this study, we utilized fixed binary skill tags as loadings, giving $a_{s_{i,t+1}} \in \{ 0,1 \}^k$, and learned $d_{s_{i,t+1}}$ during fitting. A key constraint of our implementation is that $f_Y$ takes only student latent variables and item embeddings as input, following the standard practice in most measurement functions, allowing UNVaMP to support interpretable response model configurations.

\subsection{Training Objective}
The UNVaMP training objective takes the form:
\[
    \mathcal{L} = \mathcal{L}_{\text{pred}} + \beta D_{\mathrm{KL}} ,
\]
where $\mathcal{L}_{\text{pred}}$ denotes the prediction loss (binary cross-entropy in our case), $D_{\mathrm{KL}}$ denotes the Kullback-Leibler (KL) divergence~\cite{bishop-2006-pattern}, and $\beta$ is a regularization coefficient. Because UNVaMP represents student knowledge as latent probability distributions, we require a mechanism to regularize how these distributions evolve over time. As such, we compute $D_{\mathrm{KL}}$ over temporally adjacent latent distributions:
\[
    D_{\mathrm{KL}}\left( \mathcal{N}(\mu_{i,t+1},\sigma_{i,t+1}^2) \,\|\, \mathcal{N}(\mu_{i,t},\sigma_{i,t}^2) \right) ,
\]
constraining how much the new knowledge state distribution $\mathcal{N}(\mu_{i,t+1},\sigma_{i,t+1}^2)$ diverges from the previous distribution $\mathcal{N}(\mu_{i,t},\sigma_{i,t}^2)$. This results in smoother estimated student learning dynamics and tempers the local variation of $f_z$ and $f_h$ (indirectly), with $\beta$ regulating the level of imposed smoothness. Note that $\mathcal{L}_{\text{pred}}$ and $D_{\mathrm{KL}}$ are scalar aggregated values. To prevent latent dimensionality from inflating or deflating the KL divergence, we perform mean-reduction rather than sum-reduction of these quantities. This is especially important in the case where the latent space is tied to an interpretable measurement function, since the dimensionality of the latent space should not artificially inflate or deflate the KL divergence penalty. 

While UNVaMP is inspired by the dynamic VAE literature~\cite{christie-2024-lens,girin-2021-dynamicVAEReview}, it does not perform variational Bayesian inference. Unlike standard VAE formulations, where the KL divergence is computed between an approximate posterior and a prior, UNVaMP uses the KL divergence solely as a regularization mechanism to impose smoothness on the latent knowledge distribution trajectories.

\subsection{Alignment with Architecture Desiderata}
% Dynamics regularization
% Interpretable knowledge states
% Uncertainty quantification
% Input flexibility

UNVaMP was designed to address the DLKT architecture desiderata outlined in Section~\ref{sec:desiderata}. Specifically, the $\beta$ term in UNVaMP's training objective provides a mechanism for regularizing the estimated latent dynamics, allowing users to control the trade-off between smoothness and predictive performance. UNVaMP also supports using known measurement functions (e.g., MIRT) as the response model, yielding interpretable moment-in-time knowledge state estimates and item parameters (intercepts and loadings) when so configured. In addition, UNVaMP generates latent distributions representing student knowledge states, enabling representations of internal uncertainty. Although these estimates do not provide formal coverage guarantees, they capture relative model confidence in the latent estimates. Finally, UNVaMP accommodates arbitrary response, auxiliary, and pre-response inputs. In this study, the full set of desiderata is realized under the UNVaMP-MIRT configuration with $\beta>0$, while UNVaMP-MLP retains the flexibility, uncertainty, and regularization properties without imposing interpretability constraints on the latent space.

% [Suggestion] UNVaMP was designed to address the DLKT architecture desiderata outlined in Section~\ref{sec:desiderata}. Specifically, and in the order of desirata subsections: (1) The $\beta$ term in UNVaMP's training objective provides a mechanism for regularizing the estimated latent dynamics, allowing users to control the trade-off between smoothness and predictive performance. (2) UNVaMP supports using known measurement functions (e.g., MIRT) as the response model, yielding interpretable moment-in-time knowledge state estimates when so configured. (3) UNVaMP's response model gives an internal representation of the model's uncerstainty, since it gives a distributional representation of latent traits as opposed to point values. Although these estimates do not provide formal coverage guarantees, they capture relative model confidence and, under stable data-generating conditions, tend to decrease as evidence accumulates. (4) UNVaMP accommodates arbitrary response, auxiliary, and pre-response inputs. 
% In sum, our architecture accounts for all our desired properties, and although not as rigorous as statistical KT methods, it exemplifies the type of DLKT models that would address the shortcomings discussed in Section 1.2.2 [Would need to add a label there].

\section{Methods}
% Present: properties of system and datasets
% Past: what we did

This section describes the models, datasets, and experimental procedures employed in this study. First, we introduce the UNVaMP configurations and baseline model architectures utilized in our experiments. Next, we discuss the public, internal, and simulated datasets on which our experiments were conducted. We then conclude with a description of the experimental setup and evaluation protocols.

\subsection{Models}
UNVaMP was benchmarked against a representative set of classical, hybrid, and deep learning KT models. These baselines were chosen to provide well-known comparison points with different approaches to dynamics, interpretability, and model capacity. Below, we describe the models and configurations used in our experiments.

\subsubsection{UNVaMP}

For benchmarking, we evaluated two configurations of UNVaMP that differ only in the specification of the response model. The first, UNVaMP-MLP, uses a neural network decoder as the response model. The second, UNVaMP-MIRT, is a hybrid configuration in which the response model is a 1PL MIRT measurement function. Here, the item embedder consumes pre-specified binary skill tags for each item, which are used as fixed loading parameters, while item intercepts are learned as one-dimensional embeddings. For the memory model, we used a Gated Recurrent Unit (GRU) RNN~\cite{cho-2014-gru}.

\subsubsection{BKT}
Bayesian Knowledge Tracing (BKT) is a Hidden Markov Model-based knowledge tracing method that uses response history to estimate future proficiency and correctness. The model uses four key parameters: initial knowledge probability, learning rate, guess rate (correct answer despite not knowing), and slip rate (incorrect answer despite knowing). We used the BKT implementation in the pyBKT package~\cite{badrinath2021pybkt}.

\subsubsection{Deep-IRT}
Deep-IRT is an extension of the Dynamic Key-Value Memory Network (DKVMN) architecture that predicts the probability of a correct response using a Rasch-style measurement function \cite{yeung-2019-deepIRT}. Like DKVMN, Deep-IRT employs a static key memory to compute attention weights over knowledge components, and a dynamic value memory that is updated using embeddings of student-item interactions \cite{zhang-2016-dkvmn}. Deep-IRT then estimates latent student traits and item difficulties that are combined using the Rasch measurement function to predict probability of correctness. For our experiments, we used the Deep-IRT implementation provided by the pyKT Toolkit package \cite{liu-2022-pykt}.

\subsubsection{SAINT}

SAINT is a DLKT method based on the influential transformer architecture~\cite{choi-2020-saint}. In SAINT, a multi-head attention mechanism is used within an encoder-decoder architecture, where the encoder models the exercise sequence and the decoder attends to prior responses and the encoded exercises to predict response correctness. For our experiments, we used the SAINT implementation provided by the pyKT toolkit, with minor implementation-level modifications that do not alter the model architecture. Specifically, we used PyTorch's built-in transformer encoder and decoder modules (rather than from-scratch implementations) and employed a shared positional embedding layer in the encoder and decoder blocks.

% In contrast with the original training described in \cite{choi-2020-saint}, we did not use the Noam learning rate scheduling, and simply used PyTorch's Adam optimizer with default values for the beta, weight decay, and numerical stability parameters.

\subsubsection{LKT}
We used the lasso regression implementation in the LKT package in R~\cite{lkt} to create features and train models, in which a penalty term is applied to the objective function and may reduce coefficients to zero. This approach was implemented to avoid challenges with feature selection~\cite{pavlik2023lassolkt}.

\subsection{Data}

\begin{table}[!tbp]
    \caption{Dataset summaries.}
    \centering
    \setlength{\tabcolsep}{4pt}
    \begin{tabular}{lllll}
    \toprule
    \textbf{Dataset} & \textbf{Obs.} & \textbf{Students} & \textbf{Items} & \textbf{Skills}\\
    \midrule
    ASSISTments & 172,200 & 1,561 & 548 & 78 \\
    EdNet & 2,085,026 & 18,326 & 9,198 & 187\\
    Cloze & 45,115 & 478 & 144 & 36 \\
    Amplify & 7,583,010 & 107,169 & 2,346 & 43\\
    \bottomrule
    \end{tabular}
    \label{tab:dataset-summary}
\end{table}

The data used in this study fall into three categories: public, internal, and simulated. Public and internal datasets were used for evaluating UNVaMP against existing knowledge tracing models on known benchmarks and in a real-world context, while the simulated data were used to assess UNVaMP's latent structure recovery capacity and inference behavior under controlled conditions.

\subsubsection{Public Datasets}
Three public datasets were selected for this study: ASSISTments, EdNet, and Cloze. Dataset-specific preprocessing is described below. In addition, iterative threshold filtering was applied to retain only students with responses to at least ten unique items and items with responses from at least one hundred unique students. For multi-skill items, observed per-item skill combinations were used as skill tags for all models except UNVaMP-MIRT, where latent dimensions corresponded to atomic skills. Table~\ref{tab:dataset-summary} summarizes each dataset. Each dataset was divided into a 60/20/20 train/validation/test split unless otherwise noted.

% Removed footnote URLs for space
% \footnote{\url{https://sites.google.com/view/assistmentsdatamining/dataset}}
% \footnote{\url{https://github.com/riiid/ednet}}
% \footnote{\url{https://pslcdatashop.web.cmu.edu/index.jsp?datasets=public}}
\paragraph{ASSISTments} The ASSISTments dataset contains student-item interactions from the ASSISTments online math tutoring platform, which provides immediate feedback and optional scaffolds and hints~\cite{heffernan2014assistments}. In this study, we used the ASSISTments 2017 dataset released for the 2017 ASSISTments Longitudinal Data Mining Competition~\cite{patikorn2020assistments}, a longitudinal dataset with nearly one million student actions from over 1{,}700 students spanning secondary through tertiary education. To better reflect non-intervention assessment moments, we retained only single-skill items (dropping untagged and multi-skill items) and excluded responses involving scaffolds or hints, as well as responses on student-item pairs where the initial response used assistance.

\paragraph{EdNet} The EdNet dataset contains student-item interaction data from Santa, an intelligent tutoring system used by South Korean students to prepare for the Test of English for International Communication (TOEIC)~\cite{choi2020ednet}. EdNet contains data from over 750{,}000 students and is organized into four datasets with varying levels of behavioral detail. We randomly sampled 20{,}000 students from the KT1 subset prior to threshold filtering. We then removed duplicate interactions within sessions and interactions missing responses,  imposed a cap of 500 interactions per student (approximately three standard deviations above the in-dataset mean) to reduce computational demands during model fitting.

\paragraph{Cloze} The Statistics Cloze dataset (referred to as ``Cloze'' in this study) was obtained from Memphis Datashop via the LKT R package~\cite{lkt}. The data consist of responses from Amazon Mechanical Turk participants who studied statistical concepts through short readings and completed cloze-style exercises, with each sentence associated with one of 36 skills. Sentence presentation frequency and temporal spacing (narrow, medium, wide) were experimentally manipulated, with post-practice tests administered after delays of 2 minutes, 1 day, or 3 days. We removed observations without a  correctness score. Due to the small dataset size, we used a 50/25/25 train/validation/test split with stratified sampling to maintain balance across experimental conditions.

% \paragraph{Cloze} The Statistics Cloze (referred to as \textbf{Cloze} in this study) data originally came from Memphis Datashop\footnote{https://pslcdatashop.web.cmu.edu/index.jsp?datasets=public} and were extracted from the LKT R package~\cite{lkt}. Observations come from Amazon Mechanical Turk participants who learned statistical concepts from readings and then completed exercises requiring them to fill in missing words from sentences. Each sentence is related to one of 36 skills. The presentation frequency and temporal spacing (narrow, medium, wide) of the sentences were experimentally manipulated and participants took post-practice tests after 2 minutes, 1 day, or 3 days. This experimental design was constructed to evaluate human learning patterns, such as forgetting over time and spacing effects.

% For this study, we dropped items that may have been exposed to students prior to assessment. Also, due to the small dataset size, the train/validation/test split followed a 50/25/25 ratio to ensure adequate within-set sample sizes. We employed stratified sampling to ensure balanced representation across experimental conditions. A one-way ANOVA was used to ensure that proportion correct did not vary significantly across splits. Likewise, descriptive analyses were used to confirm within-set differences between experimental groups reflected prior research. 

\subsubsection{Internal Dataset}

\paragraph{Amplify Dataset} The Amplify dataset consists of student-item interactions from a K-8 mathematics curriculum delivered via an online platform, including summative pre-unit, mid-unit, and end-of-unit assessments, as well as formative in-lesson assessments. Though responses were primarily auto-scored as correct or incorrect, when available, teacher re-scores were used in place of auto-scores. In this dataset, questions are tagged to a hierarchical proficiency map of math skills, and for this study, skill tags were drawn from the highest-level nodes. Only non-empty responses were retained and untagged items were excluded. The dataset was de-identified, with all personally identifiable information removed prior to use.

% \paragraph{Amplify Dataset} The Amplify Dataset observations came from a K-8 math curriculum built within an interactive online platform. Observations represent data from summative pre-unit, mid-unit, and end-of-unit assessments, as well as formative in-lesson assessment questions. Within these assessments, students interact with a variety of question types, most of which are automatically scored as correct or incorrect by the platform. In the platform, teachers may have the ability to rescore student work based on non auto-scored criteria (e.g., scratch work or explanations). When available, teacher re-scores were used in place of auto-scores. Items are tagged to a proficiency map, reflecting a hierarchical graph of math skills, with higher level skill nodes describing more general concepts with constituent lower level skills nested within them. For this study, skill tags were drawn from the highest level set of nodes. Only non-empty responses were retained, and items that were not tagged to skills were dropped. Additional filters were applied to remove teacher-customized content, retaining only items that were featured in the national edition of the curriculum. 

\subsubsection{Simulated Dataset}
We simulated student response data from a 1PL MIRT model. First, a global skill-skill covariance matrix $\Sigma$ was constructed with unit variances on the diagonal and off-diagonal entries drawn uniformly from the range $[0.2, 0.6]$, with $\Sigma$ constrained to be positive semi-definite. Next, a scalar overall proficiency $x_i$ was drawn from a standard normal distribution for each student. Per-student $\theta_i$ vectors were then sampled from $\mathcal{N}\left(x_i\cdot\mathbf{1}, \Sigma\right)$. Item difficulties $b_j$ were sampled from a standard normal distribution, with equal numbers of single-skill items per skill. Responses were generated by computing item intercepts $d_j=-b_j$ and applying the MIRT measurement function, then sampling Bernoulli outcomes. All students interacted with all items, and within-student item interaction order was randomized such that more difficult items tended to be seen later. Each simulated dataset contained 5 skills, 50 items per skill, and 10,000 students.

\subsection{Experiments}

\subsubsection{Model Benchmarking}
For the model benchmarking experiments, we fit all models to the public and internal datasets, then characterized predictive performance and smoothness.

Prior to benchmarking, hyperparameters for all deep learning models were tuned using Bayesian optimization. Across models, tuning included the learning rate, embedding dimensionality, and dropout probability. For UNVaMP, additional hyperparameters included the hidden state dimensionality and the number of hidden layers in the RNN and feature extractors; for UNVaMP-MLP, the number of hidden decoder layers was also tuned. For Deep-IRT, tuning included the batch size, the number of rows in the key and value memory matrices (constrained to be equal), and the maximum input sequence length. The latter was introduced to circumvent VRAM limitations, such that longer input sequences were split into subsequences and treated as additional samples during training. For SAINT, tuning included the batch size, the number of encoder and decoder layers (constrained to be equal), the number of attention heads, and the maximum input sequence length (used similarly to Deep-IRT). All DLKT models were trained using Adam-based optimizers with default settings, with UNVaMP trained using AdamW. For LKT, we included the same features as described in the original LKT-lasso paper, with the addition of power-law decay and PPE \cite{pavlik2023lassolkt}. Hyper-parameters for PPE were estimated via maximum likelihood on the training set. Other features with hyper-parameters were generated on a 0-1 grid (step=0.1).

% Hyperparameters for all deep learning models were tuned using Bayesian optimization prior to benchmarking. Across models, tuning included the learning rate, embedding dimensionality, and dropout probability. For UNVaMP, additional hyperparameters included: hidden state dimensionality, number of hidden layers in the RNN, and number of hidden layers in the feature extractors; and, for UNVaMP-MLP, the number of hidden decoder layers. For Deep IRT, tuning included the batch size, the number of rows in the key and value memory matrices (constrained to be equal), and the maximum input sequence length. For SAINT, tuning included the batch size, the number of encoder and decoder layers (constrained to be equal), the number of attention heads, and the maximum input sequence length. All benchmark models were trained using Adam-based optimizers with default settings, with UNVaMP trained using AdamW.

% In addition, ASSISTments included problem type indicators, EdNet included bundle identifiers, and the Amplify dataset included item type indicators (e.g., multiple choice) and other item metadata. For Cloze, only submission timestamps were used as auxiliary inputs. 
Because LKT and UNVaMP allow flexible specification of input features, we evaluated two versions of each model. The minimal versions were trained using the same inputs as the other benchmark models: item ID, skill ID, sequence position, and response correctness (and features derived from these inputs). The (aux) versions included all minimal inputs plus available auxiliary inputs, which varied by dataset (e.g., problem type, bundle ID). All datasets provided submission timestamps, from which recency features (e.g., time since previous submission) were computed.

% Benchmark fits of UNVaMP-MLP and UNVaMP-MIRT were conducted with $\beta \in \{0, 10^{-3}\}$. These values of $\beta$ were selected to demonstrate UNVaMP's peak predictive performance, as larger values of $\beta$ introduced a trade-off between regularization and predictive performance. 

Benchmark fits of UNVaMP-MLP and UNVaMP-MIRT were conducted using $\beta \in \{0, 10^{-3}\}$ to reflect peak predictive performance. To examine the trade-off between predictive performance and regularization, we compared UNVaMP-MIRT (aux) fits across $\beta \in \{0, 10^{-3}, 10^{-2}, 10^{-1}\}$ for each dataset. To quantify the smoothing effect of $\beta$ on the latent dynamics, we report what we call the \textit{average latent total variation} (ALTV). ALTV can be viewed as a normalized, fixed-sample analogue of \textit{total variation}, a concept in real analysis that describes the oscillatory behavior of functions. This metric was also inspired by the $w_1$ smoothness metric described in~\cite{yeung-2018-dktPlus}, though ALTV is computed in the latent space and gives equal weight to each student. ALTV is formulated as
\[
    \text{ALTV}=\frac{1}{N}\sum_{i=1}^{N} \left(
        \frac{1}{k(T_i-1)} \sum_{t=1}^{T_i-1}\lVert z_{i,t+1} - z_{i,t} \rVert_{1}
    \right),
\]
where $z_{i,t} \in \mathbb{R}^k$ is the latent proficiency vector for student $i\in\{1,\cdots,N\}$ at time point $t$. Under the 1PL MIRT measurement function, each latent dimension enters the logit with a binary coefficient. Consequently, holding all else fixed, a one unit perturbation to $z_{i,t}$ in any latent dimension has the same effect on the logit. This means latent dimensions share units and the use of the $L^1$ norm over the differences $(z_{i,t+1} - z_{i,t})$ is justified.

\subsubsection{Simulation Study}

In the simulation study, we evaluated UNVaMP-MIRT under five levels of Gaussian measurement noise, perturbing each student response using
\[
    P\left[ Y_{i,j}=1 \mid \theta_i, a_j, d_j \right] = \frac{1}{1 + \exp\{-(a_j^\top \theta_i + d_j + \epsilon)\}} ,
\]
where $\epsilon \sim \mathcal{N}(0, \nu)$ and $\nu \in \{0.0, 0.25, 0.5, 0.75, 1.0\}$. For each noise level, 5 replications were conducted (25 total runs). The skill-skill covariance matrix was fixed across runs, with student proficiencies, item difficulties, and responses re-sampled within each run. UNVaMP-MIRT and baseline MIRT models were fit separately within each run, with MIRT estimates obtained via maximum a posteriori (MAP) using a standard normal prior. MAP was chosen over maximum likelihood estimation (MLE) or weighted maximum likelihood estimation (WLE) for numerical stability and reliable convergence. Both models were fit with ground-truth item parameters $a_j$ and $d_j$ fixed, aligning latent scales such that student latent variables were the only estimated quantities. We examined two primary behaviors of UNVaMP-MIRT: (1) latent variable recovery accuracy, and (2) the behavior of uncertainty estimates as a function of the number of observations. Each replication comprised 10,000 students with 250 responses per student (2.5M observations per run). Except where otherwise noted, all simulation analyses use UNVaMP-MIRT with $\beta = 10^{-2}$.

\section{Results}

\subsection{Model Benchmarking}

\begin{table*}[!tbp]
\centering
\caption{Model performance across four real-world datasets. Best values per dataset, per metric are in bold.}
\setlength{\tabcolsep}{4pt}
\renewcommand{\arraystretch}{1.2}
\begin{tabular}{lcccccccc}
\toprule
& \multicolumn{2}{c}{\textbf{ASSISTments}} 
& \multicolumn{2}{c}{\textbf{EdNet}} 
& \multicolumn{2}{c}{\textbf{Cloze}} 
& \multicolumn{2}{c}{\textbf{Amplify}} \\
\cmidrule(lr){2-3} \cmidrule(lr){4-5} \cmidrule(lr){6-7} \cmidrule(lr){8-9}
\textbf{Model} & AUC & Acc.
& AUC & Acc.
& AUC & Acc. 
& AUC & Acc. \\
\midrule
BKT             & 0.635 & 0.595
                & 0.668 & 0.659 
                & 0.724 & 0.674
                & 0.654 & 0.614 \\
Deep-IRT        & 0.687 & 0.633
                & 0.697 & 0.675
                & 0.753 & 0.689
                & 0.768 & 0.697 \\
SAINT           & 0.698 & 0.639 
                & 0.748 & 0.704
                & 0.673 & 0.625
                & 0.868 & \textbf{0.780} \\
Lasso LKT       & 0.732 & 0.665
                & 0.761 & 0.713
                & 0.846 & 0.766
                & 0.810 & 0.734 \\
Lasso LKT (aux) & 0.736 & 0.670
                & \textbf{0.767} & 0.716
                & \textbf{0.858} & \textbf{0.775}
                & 0.822 & 0.745 \\
\midrule
UNVaMP-MLP ($\beta=0$)        & 0.743 & 0.677 
                              & \textbf{0.767} & \textbf{0.717}
                              & 0.833 & 0.750
                              & 0.867 & 0.779 \\
UNVaMP-MLP ($\beta=10^{-3}$)  & 0.742 & 0.675
                              & 0.766 & \textbf{0.717}
                              & 0.827 & 0.740
                              & \textbf{0.869} & \textbf{0.780} \\
UNVaMP-MLP (aux, $\beta=0$)        & \textbf{0.745} & 0.677
                                   & 0.765 & 0.716
                                   & 0.837 & 0.756
                                   & 0.865 & 0.777 \\
UNVaMP-MLP (aux, $\beta=10^{-3}$)  & \textbf{0.745} & \textbf{0.678}
                                   & 0.765 & 0.716
                                   & 0.831 & 0.751
                                   & 0.865 & 0.776 \\

UNVaMP-MIRT ($\beta=0$)       & 0.730 & 0.666
                              & 0.762 & 0.714
                              & 0.821 & 0.742
                              & 0.844 & 0.759 \\
UNVaMP-MIRT ($\beta=10^{-3}$) & 0.729 & 0.662
                              & 0.755 & 0.709
                              & 0.810 & 0.737
                              & 0.845 & 0.760 \\
UNVaMP-MIRT (aux, $\beta=0$)       & 0.736 & 0.673
                                   & 0.755 & 0.709
                                   & 0.815 & 0.738
                                   & 0.849 & 0.764 \\
UNVaMP-MIRT (aux, $\beta=10^{-3}$) & 0.738 & 0.670
                                   & 0.751 & 0.706
                                   & 0.803 & 0.734
                                   & 0.850 & 0.764 \\

\bottomrule
\end{tabular}
\label{tab:benchmark-results}
\end{table*}

\begin{figure}[!tbp]
    \centering
    \includegraphics[width=3.25in]{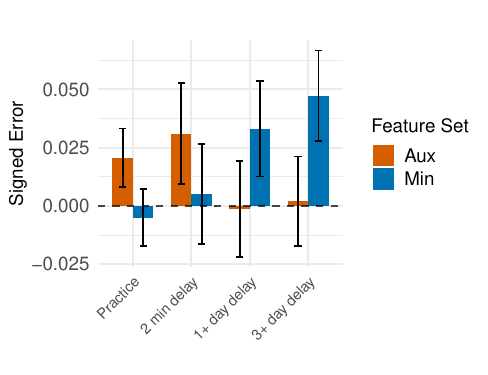}
    \caption{
        UNVaMP-MIRT signed error ($\text{mean}(\hat{p} - y)$) on the Cloze dataset with and without auxiliary inputs, under $\beta=10^{-3}$. With auxiliary features included, UNVaMP-MIRT demonstrates reduced error in delayed post-tests. Error bars represent 95\% confidence intervals. Standard errors were computed as the SD of prediction residuals divided by the square root of the subset sample size (practice session, 2 minute delay, 1+ day delay, or 3+ day delay).
    }
    \label{fig:cloze-min-aux-error}
    \Description{
        This figure shows signed prediction error on the Cloze dataset across four experimental conditions: practice, 2 minute delay, 1+ day delay, and 3+ day delay. The horizontal axis lists the conditions, and the vertical axis shows signed error, where positive values indicate over-prediction and negative values indicate under-prediction.
    
        For each condition, results are shown for two feature sets: a minimal feature set and an auxiliary feature set. Bars represent mean signed error, and vertical error bars indicate uncertainty around the estimates. In the practice condition, both feature sets produce signed errors near zero, with the auxiliary set slightly overestimating and the minimal set slightly underestimating, but near zero. With a 2 minute delay, the auxiliary set overestimates slightly more than before, while the minimal set slightly overestimates but is near zero. In the delayed conditions, the minimal feature set shows increasingly positive signed error as delay increases, indicating growing overprediction. In contrast, the auxiliary feature set is roughly zero across 1+ day and 3+ day delay conditions, showing reduced bias.
    }
\end{figure}

\begin{table*}[!tbp]
\centering
\caption{Impact of increasing $\beta$ on AUC and ALTV. Highest AUC and lowest ALTV values per dataset are in bold.}
\setlength{\tabcolsep}{4pt}
\renewcommand{\arraystretch}{1.2}
\begin{tabular}{lcccccccc}
\toprule
& \multicolumn{2}{c}{\textbf{ASSISTments}} 
& \multicolumn{2}{c}{\textbf{EdNet}} 
& \multicolumn{2}{c}{\textbf{Cloze}} 
& \multicolumn{2}{c}{\textbf{Amplify}} \\
\cmidrule(lr){2-3} \cmidrule(lr){4-5} \cmidrule(lr){6-7} \cmidrule(lr){8-9}
\textbf{Model} & AUC & ALTV
& AUC & ALTV
& AUC & ALTV 
& AUC & ALTV \\
\midrule

UNVaMP-MIRT (aux, $\beta=0$)       & 0.736 & 0.418
                                   & \textbf{0.755} & 0.185
                                   & \textbf{0.815} & 0.322
                                   & 0.849 & 0.765 \\
UNVaMP-MIRT (aux, $\beta=10^{-3}$) & \textbf{0.738} & 0.039
                                   & 0.751 & 0.017
                                   & 0.803 & 0.209
                                   & \textbf{0.850} & 0.192 \\
UNVaMP-MIRT (aux, $\beta=10^{-2}$) & 0.722 & 0.015
                                   & 0.747 & 0.012
                                   & 0.733 & 0.039
                                   & 0.844 & 0.072 \\
UNVaMP-MIRT (aux, $\beta=10^{-1}$) & 0.703 & \textbf{0.007}
                                   & 0.744 & \textbf{0.009}
                                   & 0.713 & \textbf{0.017}
                                   & 0.802 & \textbf{0.023} \\

% % With Acc. column
% UNVaMP-MIRT (aux, $\beta=0$)       & 0.736 & \textbf{0.673} & -
%                                    & \textbf{0.755} & \textbf{0.709} & -
%                                    & \textbf{0.815} & \textbf{0.738} & -
%                                    & 0.849 & \textbf{0.764} & - \\
% UNVaMP-MIRT (aux, $\beta=10^{-3}$) & \textbf{0.738} & 0.670 & -
%                                    & 0.751 & 0.706 & -
%                                    & 0.803 & 0.734 & -
%                                    & \textbf{0.850} & \textbf{0.764} & - \\
% UNVaMP-MIRT (aux, $\beta=10^{-2}$) & 0.722 & 0.656 & -
%                                    & 0.747 & 0.703 & -
%                                    & 0.733 & 0.667 & -
%                                    & 0.844 & 0.760 & - \\
% UNVaMP-MIRT (aux, $\beta=10^{-1}$) & 0.703 & 0.642 & -
%                                    & 0.744 & 0.702 & -
%                                    & 0.713 & 0.662 & -
%                                    & 0.802 & 0.728 & - \\

\bottomrule
\end{tabular}
\label{tab:beta-benchmark-results}
\end{table*}

UNVaMP-MLP achieved the strongest overall performance during model benchmarking, outperforming or matching all other models on three of four datasets in both AUC and accuracy, with LKT performing best on the Cloze dataset (Table~\ref{tab:benchmark-results}). UNVaMP-MIRT performed competitively, with only modest losses in predictive performance relative to UNVaMP-MLP, indicating that the cost of an interpretable response model is small in practice. Specifically, UNVaMP-MIRT outperformed SAINT on ASSISTments, EdNet, and Cloze, and performed slightly worse on the Amplify dataset. Compared to LKT, UNVaMP-MIRT performed comparably on ASSISTments and EdNet, and better on the Amplify dataset, though worse on Cloze. Both UNVaMP variants showed strong performance across datasets, despite varying dataset sizes, item and skill counts, and other structural attributes, indicating robustness to dataset-specific characteristics. By contrast, SAINT's performance appeared to be highly sensitive to dataset size, consistent with its transformer-based architecture. Differences between our baseline model results and those reported in the original works are primarily attributable to our data preprocessing and subsampling choices.

% In the particular case of EdNet, we keep about 2.85\% of the original number of observations in this dataset, leading to lower AUC and accuracy than the ones originally reported \cite{choi-2020-saint}.

LKT's strong performance on Cloze is unsurprising, given that the data were collected under constrained, experimental conditions, with relatively few students and observations, limiting the performance of higher-capacity DLKT methods. LKT also benefited from hand-designed, research-based features, which can outperform deep learning methods in low- and medium-data settings~\cite{gervet2020deep}. Furthermore, Table~\ref{tab:benchmark-results} shows that the impact of auxiliary inputs on UNVaMP's aggregate predictive performance was minimal, with their inclusion producing negligible performance differences for UNVaMP-MLP and small, mixed effects for UNVaMP-MIRT.

Although auxiliary features had a limited impact on aggregate predictive performance for UNVaMP, results for the Cloze dataset---with its spacing, repetition, and retention interval manipulations---indicate that these features meaningfully influenced model behavior. Stratifying UNVaMP-MIRT performance by post-test session interval revealed that longer retention intervals were associated with smaller signed prediction errors for the auxiliary configuration (Fig.~\ref{fig:cloze-min-aux-error}). This suggests that auxiliary features (e.g., elapsed time) can improve sensitivity to meaningful underlying structure, despite modest losses in aggregate accuracy. We suspect this sensitivity would improve, and perhaps come at reduced predictive cost (if any), given a larger dataset.

% Although auxiliary features had limited impact on overall predictive performance, there are indications that they meaningfully influenced model behavior within UNVaMP. The Cloze dataset provides a useful case study due to its experimental design and interpretable condition structure, including within-student manipulations of spacing and repetition, and between-student manipulations of retention interval (2 minutes, 1+ day, 3+ days). Overall, UNVaMP-MIRT with auxiliary features (e.g., elapsed time between attempts) performed slightly worse than the minimal configuration in terms of aggregate AUC and accuracy. However, stratifying performance by posttest session interval revealed a consistent pattern: longer delays were associated with smaller signed prediction errors for the model with auxiliary features (Fig.~\ref{fig:cloze-min-aux-error}). This suggests that despite modest losses in aggregate accuracy, including auxiliary features can improve sensitivity to meaningful covariates and underlying structure. We suspect that the benefits of including auxiliary inputs would have been more pronounced had the Cloze dataset contained more students and observations.

To examine the effect of $\beta$ on predictive performance, we evaluated UNVaMP-MIRT (aux) under the conditions $\beta \in \{ 0, 10^{-3}, 10^{-2}, 10^{-1} \}$, revealing a clear pattern (Table~\ref{tab:beta-benchmark-results}). Increasing $\beta$ from 0 to $10^{-3}$ produced no meaningful change in prediction accuracy, whereas the change from $10^{-3}$ to $10^{-2}$ led to a small but consistent decrease across datasets. A further increase to $10^{-1}$ resulted in a substantial accuracy drop in most cases. These results indicate that small values of $\beta$ have little practical impact on predictive performance. Across all datasets, ALTV decreased monotonically with increasing $\beta$, confirming that $\beta$ provides control over the smoothness of latent trajectories. With the exception of Cloze, the increase in $\beta$ from 0 to $10^{-3}$ resulted in the largest decrease in ALTV by far. Subsequent $\beta$ increases showed diminishing returns, along with increased performance penalties. The smoothing effects of $\beta$ are further examined qualitatively in the simulation study results below.

\subsection{Simulation Study}

\begin{figure*}[!tbp]
    \centering    
    \includegraphics[width=5.5in]{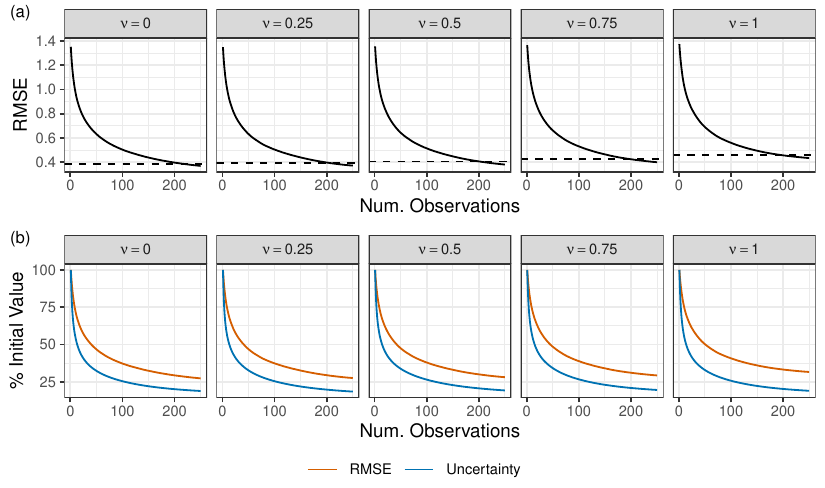}
    \caption{
        Relationship between the fit metrics of interest and the number of observations (student-item interactions).
        Panel \textbf{(a)} shows RMSE vs. observation count, averaged across replications. Cross-replication SD is omitted from the figure due to small magnitude (averaging below $0.01$ for all noise conditions). The dashed line indicates baseline MIRT final RMSE, computed after observing all data. 
        Panel \textbf{(b)} shows the percent of the initial value (RMSE and uncertainty) vs. observation count, averaged across replications. Again, cross-replication SD is omitted from the figure due to small magnitudes (averaging below $1\%$ for RMSE and below $3\%$ for uncertainty across all noise conditions).
    }
    \label{fig:rmse-combined}
    \Description{
        This figure shows how estimation error and uncertainty change as the number of student–item observations increases in the simulation study, across multiple noise conditions. The figure contains two rows of panels, each faceted by noise level, with noise values ranging from zero to one.

        Panel (a) displays root mean squared error (RMSE) as a function of the number of observations. For all noise levels, RMSE decreases rapidly as more observations are observed and then gradually levels off. A dashed horizontal line indicates the baseline RMSE obtained from fitting a standard MIRT model using all available data. Across noise conditions, UNVaMP’s RMSE approaches this baseline as the number of observations increases.
        
        Panel (b) shows RMSE and latent uncertainty expressed as a percentage of their initial values, plotted against the number of observations. Both quantities decrease monotonically with additional observations, with uncertainty declining more quickly than RMSE in the early stages (around observations 20-30). This pattern is consistent across all noise levels, indicating that the model becomes more confident and more accurate as evidence accumulates.

    }

\end{figure*}

\begin{figure}[!tbp]
\centering
\includegraphics[width=3.25in]{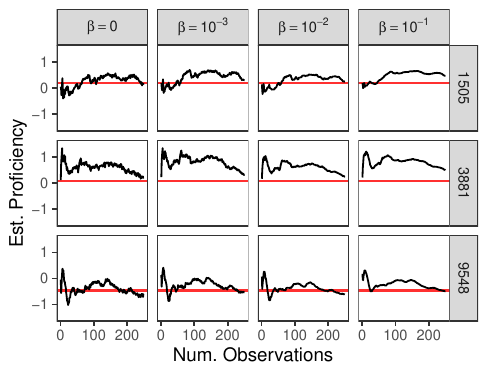}
\caption{
    Effect of $\beta$ on latent proficiency estimation. Columns correspond to $\beta \in \{0, 10^{-3}, 10^{-2}, 10^{-1}\}$; rows correspond to individual simulated students. The black line shows the UNVaMP-MIRT estimate for one latent skill plotted against the cumulative number of responses across all skills. The red horizontal line denotes the true value of the selected skill.
}
\label{fig:latents-and-beta}
\Description{
    This figure illustrates the effect of the KL regularization parameter beta on the volatility of latent proficiency estimates in a simulation study. The figure is arranged as a grid, with columns corresponding to increasing values of beta and rows corresponding to different simulated students.
    
    Each panel shows the estimated proficiency for a single latent skill plotted against the cumulative number of observations. The black line represents the UNVaMP-MIRT estimated proficiency trajectory over time, while a horizontal red line indicates the true latent proficiency value used to generate the data.
    
    When beta is zero or very small, estimated proficiency trajectories exhibit higher short-term variability and fluctuate more around the true value. As beta increases, trajectories become smoother, with estimates exhibiting reduced fluctuation around the true value. Across students, higher beta values consistently dampen abrupt changes in estimated proficiency, demonstrating that the regularization parameter controls the smoothness of latent knowledge estimation.
}
\end{figure}

In the simulation study, we evaluated UNVaMP-MIRT's ability to recover student latent variables relative to a baseline MIRT model. Fig.~\ref{fig:rmse-combined}(a) shows that, across all noise conditions, UNVaMP-MIRT performed comparably to the MIRT baseline once it had observed the full response sequence. Given the large number of responses per student, MAP shrinkage in the MIRT baseline model is unlikely to have materially affected accuracy. These results do not indicate that UNVaMP-MIRT will outperform MIRT under idealized conditions, but rather that it is capable of recovering latent variables with comparable accuracy given sufficient data. UNVaMP-MIRT's competitive recovery accuracy in this experiment may partly reflect its ability to implicitly exploit the cross-skill correlations that were built into the simulation process, whereas the baseline MIRT models were fit under a simple per-skill standard normal prior.

% In the simulation study, we evaluated UNVaMP-MIRT's ability to recover student latent variables relative to a baseline MIRT model. Fig.~\ref{fig:rmse-combined}(a) shows recovery RMSE for UNVaMP-MIRT after observing $1,2,\dots,250$ responses, alongside the RMSE obtained by the MIRT baseline. Across all noise conditions, UNVaMP-MIRT achieved RMSE comparable to the MIRT baseline once it had observed the full response sequence. Given the large number of responses per student, MAP shrinkage in the MIRT baseline is unlikely to have materially affected recovery accuracy. We make no claims that UNVaMP-MIRT outperforms MIRT under idealized simulated conditions; rather, the results indicate that with sufficient data, UNVaMP recovers latent variables at a level of accuracy very close to a baseline MIRT model. The fact that the data were simulated to have plausible skill-skill correlations (between $0.2$ and $0.6$) is a possible contributor to UNVaMP-MIRT's competitive recovery accuracy, since it can implicitly exploit cross-skill structure during sequential inference. The baseline MIRT model can incorporate such correlation structure only through an explicit prior, and our experiments utilized a simple per-skill standard normal. Cross-replication RMSE variability was low for UNVaMP-MIRT (averaging below 0.01 for all noise conditions), indicating stable recovery behavior.

Fig.~\ref{fig:rmse-combined}(b) shows that UNVaMP-MIRT's uncertainty estimates decreased alongside recovery RMSE across noise conditions, indicating that model uncertainty tracked recovery accuracy. Uncertainty declined more rapidly than RMSE early in the sequence, suggesting some possible early overconfidence, but both quantities exhibited a shared monotonic decrease. Fig.~\ref{fig:latents-and-beta} illustrates the effect of $\beta$ on latent variable estimation, with larger values producing less volatile estimation behavior. We emphasize that this simulation study considered static ground-truth latent knowledge states; further analysis is needed to evaluate model behavior under dynamically evolving states.

% Fig.  \ref{fig:rmse-combined}(b) shows the evolution of recovery RMSE and estimated uncertainty for UNVaMP-MIRT, plotted as a percentage of their initial values after each observation. Across noise conditions, uncertainty estimates decreased alongside recovery error, indicating that model uncertainty tracked latent recovery accuracy. Although uncertainty declined more rapidly than RMSE early in the sequence, both quantities exhibited a shared monotonic reduction, suggesting internally consistent uncertainty adjustment, despite some possible early overconfidence. Fig.\ref{fig:latents-and-beta} illustrates the effect of $\beta$ regularization on student latent variable estimation, with larger $\beta$ values resulting in less volatile estimation behavior. We emphasize that this simulation study considered static ground-truth latent knowledge states; further analysis is needed to evaluate model behavior under dynamically evolving latent knowledge states.

\section{Discussion}
In total, our findings show that UNVaMP offers practical mechanisms for regularizing latent dynamics and modeling latent structure within student-item interaction data, all while maintaining competitive predictive performance. Below, we clarify the scope and limitations of the current work and place these results in a broader context.

\subsection{Limitations and Future Work}

Although ALTV quantifies the effect of $\beta$ on latent trajectory smoothness, connecting such metrics to practical criteria for $\beta$ selection is not straightforward. For instance, student knowledge estimates are expected to exhibit high early volatility, due to both estimation noise and genuinely rapid early learning, followed by more moderated trajectories. It remains unclear how to weigh smoothness against predictive accuracy, or whether such measures would have an intuitive interpretation. These issues arise because, although a hybrid UNVaMP configuration offers interpretable snapshots of estimated student knowledge, the underlying dynamics are still the product of a deep learning model and are not governed by interpretable updating rules.

Likewise, while UNVaMP provides quantitative uncertainty estimates, those estimates are not calibrated according to any interpretable statistical framework and should not be used to provide coverage guarantees; rather, they are best used to identify changes in relative confidence within the model as evidence accumulates. Because our current study examined the relationship between UNVaMP's latent uncertainty estimates and recovery accuracy only under simulated conditions, the relationship between uncertainty and external signals (e.g., predictive accuracy) should be investigated in a real-world setting; however, such investigations would need to account for the sparsity, noise, and heterogeneity of real-world interaction data.

UNVaMP can utilize arbitrary engineered features as auxiliary inputs, such as those used in LKT models. Future work could examine the impact of incorporating these features on both predictive performance and interpretability. In addition, subsequent studies could investigate UNVaMP's ability to implicitly recover experimentally demonstrated learning phenomena from data in which such structure is known to be present. While experimentally generated datasets such as Cloze are valuable for this purpose, their limited size constrains the direct application of DLKT methods. Accordingly, future work could include large experimentally informed simulation studies to examine how auxiliary inputs influence UNVaMP's sensitivity to known learning phenomena.

\subsection{Conclusions}
In this paper, we introduced UNVaMP, a neural network knowledge tracing architecture that estimates latent distributions over student knowledge as it evolves over time. We found that a pure neural configuration (UNVaMP-MLP) achieved the strongest predictive performance among compared models on three out of four datasets, while a hybrid configuration (UNVaMP-MIRT) lagged only slightly behind, highlighting the relatively small penalty incurred for improving interpretability. In addition, UNVaMP-MIRT with auxiliary inputs exhibited sensitivity to underlying temporal structure in the experimental Cloze dataset. Under simulated conditions, the UNVaMP-MIRT configuration further exhibited robust latent variable recovery and internally consistent uncertainty behavior. Taken together, these results indicate that UNVaMP, particularly in its hybrid configuration, is a practical solution for applications that require not only strong predictive performance, but also regularized latent knowledge trajectories, interpretable knowledge state estimates, internally consistent uncertainty quantification, and flexible input modeling.

\bibliographystyle{abbrv}
\bibliography{refs}

\begin{thebibliography}{10}

\bibitem{badrinath2021pybkt}
A.~Badrinath, F.~Wang, and Z.~Pardos.
\newblock pybkt: An accessible python library of bayesian knowledge tracing models, 2021.
\newblock arXiv:2105.00385.

\bibitem{baker-2008-bkt}
R.~S. J.~d. Baker, A.~T. Corbett, and V.~Aleven.
\newblock More accurate student modeling through contextual estimation of slip and guess probabilities in {Bayesian} knowledge tracing.
\newblock In B.~P. Woolf, E.~Aïmeur, R.~Nkambou, and S.~Lajoie, editors, {\em Intelligent {Tutoring} {Systems}}, pages 406--415, Berlin, Heidelberg, 2008. Springer.

\bibitem{bishop-2006-pattern}
C.~M. Bishop.
\newblock {\em Pattern recognition and machine learning}.
\newblock Springer, 2006.

\bibitem{borsboom2008latent}
D.~Borsboom.
\newblock Latent variable theory.
\newblock {\em Measurement}, 6:25--53, 2008.

\bibitem{cen2006AFM1}
H.~Cen, K.~Koedinger, and B.~Junker.
\newblock Learning factors analysis--a general method for cognitive model evaluation and improvement.
\newblock In {\em International conference on intelligent tutoring systems}, pages 164--175. Springer, 2006.

\bibitem{chen-2023-qikt}
J.~Chen, Z.~Liu, S.~Huang, Q.~Liu, and W.~Luo.
\newblock Improving interpretability of deep sequential knowledge tracing models with question-centric cognitive representations, 2023.
\newblock arXiv:2302.06885.

\bibitem{chen-2018-dynamicCDM}
Y.~Chen, S.~A. Culpepper, S.~Wang, and J.~Douglas.
\newblock A hidden markov model for learning trajectories in cognitive diagnosis with application to spatial rotation skills.
\newblock {\em Applied Psychological Measurement}, 42(1):5--23, Jan. 2018.

\bibitem{cheng-2025-uncertaintyAwareKT}
W.~Cheng, H.~Du, C.~Li, E.~Ni, L.~Tan, T.~Xu, and Y.~Ni.
\newblock Uncertainty-aware knowledge tracing, 2025.
\newblock arXiv:2501.05415.

\bibitem{cho-2014-gru}
K.~Cho, B.~van Merrienboer, D.~Bahdanau, and Y.~Bengio.
\newblock On the properties of neural machine translation: Encoder-decoder approaches, 2014.
\newblock arXiv:1409.1259.

\bibitem{choi-2020-saint}
Y.~Choi, Y.~Lee, J.~Cho, J.~Baek, B.~Kim, Y.~Cha, D.~Shin, C.~Bae, and J.~Heo.
\newblock Towards an appropriate query, key, and value computation for knowledge tracing.
\newblock In {\em {LAK21}: 11th {International} {Learning} {Analytics} and {Knowledge} {Conference}}, pages 490--496, Apr. 2020.

\bibitem{choi2020ednet}
Y.~Choi, Y.~Lee, D.~Shin, J.~Cho, S.~Park, S.~Lee, J.~Baek, C.~Bae, B.~Kim, and J.~Heo.
\newblock Ednet: A large-scale hierarchical dataset in education.
\newblock In {\em International conference on artificial intelligence in education}, pages 69--73. Springer, 2020.

\bibitem{christie-2024-lens}
S.~T. Christie, C.~Cook, and A.~N. Rafferty.
\newblock Uncertainty-preserving deep knowledge tracing with state-space models, 2024.
\newblock arXiv:2407.17427.

\bibitem{corbett-1995-bkt}
A.~T. Corbett and J.~R. Anderson.
\newblock Knowledge tracing: {Modeling} the acquisition of procedural knowledge.
\newblock {\em User Modelling and User-Adapted Interaction}, 4(4):253--278, 1995.

\bibitem{galyardt2015RPFA}
A.~Galyardt and I.~Goldin.
\newblock Move your lamp post: Recent data reflects learner knowledge better than older data.
\newblock {\em Journal of Educational Data Mining}, 7(2):83--108, 2015.

\bibitem{gervet2020deep}
T.~Gervet, K.~Koedinger, J.~Schneider, T.~Mitchell, et~al.
\newblock When is deep learning the best approach to knowledge tracing?
\newblock {\em Journal of Educational Data Mining}, 12(3):31--54, 2020.

\bibitem{ghosh-2020-akt}
A.~Ghosh, N.~Heffernan, and A.~S. Lan.
\newblock Context-aware attentive knowledge tracing, 2020.
\newblock arXiv:2007.12324.

\bibitem{girin-2021-dynamicVAEReview}
L.~Girin, S.~Leglaive, X.~Bie, J.~Diard, T.~Hueber, and X.~Alameda-Pineda.
\newblock Dynamical variational autoencoders: A comprehensive review.
\newblock {\em Foundations and Trends in Machine Learning}, 15(1–2):1–175, 2021.

\bibitem{gong2010comparing}
Y.~Gong, J.~E. Beck, and N.~T. Heffernan.
\newblock Comparing knowledge tracing and performance factor analysis by using multiple model fitting procedures.
\newblock In {\em International conference on intelligent tutoring systems}, pages 35--44. Springer, 2010.

\bibitem{gonzalez-2014-elo}
J.~González-Brenes, Y.~Huang, and P.~Brusilovsky.
\newblock General features in knowledge tracing: Applications to multiple subskills, temporal item response theory, and expert knowledge.
\newblock In {\em Proceedings of the 7th International Conference on Educational Data Mining}, 2014.

\bibitem{hambleton1991fundamentals_irt}
R.~K. Hambleton, H.~Swaminathan, and H.~J. Rogers.
\newblock {\em Fundamentals of item response theory}, volume~2.
\newblock Sage, 1991.

\bibitem{heffernan2014assistments}
N.~T. Heffernan and C.~L. Heffernan.
\newblock The assistments ecosystem: Building a platform that brings scientists and teachers together for minimally invasive research on human learning and teaching.
\newblock {\em International Journal of Artificial Intelligence in Education}, 24(4):470--497, 2014.

\bibitem{kaya-2016-dynamicCDM}
Y.~Kaya and W.~L. Leite.
\newblock Assessing change in latent skills across time with longitudinal cognitive diagnosis modeling: {An} evaluation of model performance.
\newblock {\em Educational and Psychological Measurement}, 77(3):369--388, June 2016.

\bibitem{khajah-2016-howdeepiskt}
M.~Khajah, R.~V. Lindsey, and M.~C. Mozer.
\newblock How deep is knowledge tracing?, 2016.
\newblock arXiv:1604.02416.

\bibitem{lee-2025-dynamicIRT}
H.~Lee, J.~B. Cho, D.~S. Matteson, and B.~W. Domingue.
\newblock Dynamic bayesian item response model with decomposition ({D-BIRD}): {Modeling} cohort and individual learning over time, 2025.
\newblock arXiv:2506.21723.

\bibitem{lindsey2014colt}
R.~V. Lindsey, J.~D. Shroyer, H.~Pashler, and M.~C. Mozer.
\newblock Improving students’ long-term knowledge retention through personalized review.
\newblock {\em Psychological science}, 25(3):639--647, 2014.

\bibitem{liu-2022-pykt}
Z.~Liu, Q.~Liu, J.~Chen, S.~Huang, J.~Tang, and W.~Luo.
\newblock py{KT}: A python library to benchmark deep learning based knowledge tracing models.
\newblock In {\em Thirty-sixth Conference on Neural Information Processing Systems Datasets and Benchmarks Track}, 2022.

\bibitem{pandey-2019-sakt}
S.~Pandey and G.~Karypis.
\newblock A self-attentive model for knowledge tracing, 2019.
\newblock arXiv:1907.06837.

\bibitem{pandey-2020-rkt}
S.~Pandey and J.~Srivastava.
\newblock {RKT}: Relation-aware self-attention for knowledge tracing.
\newblock In {\em Proceedings of the 29th ACM International Conference on Information \& Knowledge Management}, CIKM ’20, page 1205–1214. ACM, Oct. 2020.

\bibitem{patikorn2020assistments}
T.~Patikorn, R.~S. Baker, N.~T. Heffernan, et~al.
\newblock Assistments longitudinal data mining competition special issue: a preface.
\newblock {\em Journal of Educational Data Mining}, 12(2):i--xi, 2020.

\bibitem{pavlik2008optimalspacing}
P.~I. Pavlik and J.~R. Anderson.
\newblock Using a model to compute the optimal schedule of practice.
\newblock {\em Journal of Experimental Psychology: Applied}, 14(2):101--117, 2008.

\bibitem{pavlik2009PFA}
P.~I. Pavlik, H.~Cen, and K.~R. Koedinger.
\newblock Performance factors analysis -- a new alternative to knowledge tracing.
\newblock In {\em Proceedings of the 2009 Conference on Artificial Intelligence in Education: Building Learning Systems That Care: From Knowledge Representation to Affective Modelling}, page 531–538, NLD, 2009. IOS Press.

\bibitem{pavlik2021LKTpaper}
P.~I. Pavlik, L.~G. Eglington, and L.~M. Harrell-Williams.
\newblock Logistic knowledge tracing: A constrained framework for learner modeling.
\newblock {\em IEEE Transactions on Learning Technologies}, 14(5):624--639, 2021.

\bibitem{lkt}
P.~I. {Pavlik Jr.} and L.~G. Eglington.
\newblock {\em LKT: Logistic Knowledge Tracing}, 2024.
\newblock R package version 1.7.0.

\bibitem{pavlik2025evolutionary}
P.~I. Pavlik~Jr and L.~G. Eglington.
\newblock Evolutionary features for mitigating cold starts in logistic knowledge tracing.
\newblock {\em International Educational Data Mining Society}, 2025.

\bibitem{pavlik2023lassolkt}
P.~I. Pavlik~Jr, L.~G. Eglington, et~al.
\newblock Automated search improves logistic knowledge tracing, surpassing deep learning in accuracy and explainability.
\newblock {\em Journal of Educational Data Mining}, 15(3):58--86, 2023.

\bibitem{pelanek-2016-elo}
R.~Pelánek.
\newblock Applications of the {Elo} rating system in adaptive educational systems.
\newblock {\em Computers \& Education}, 98:169--179, July 2016.

\bibitem{piech-dkt-2015}
C.~Piech, J.~Spencer, J.~Huang, S.~Ganguli, M.~Sahami, L.~Guibas, and J.~Sohl-Dickstein.
\newblock Deep knowledge tracing, 2015.
\newblock arXiv:1506.05908.

\bibitem{rachatasumrit2024interpretability}
N.~Rachatasumrit, P.~Carvalho, and K.~Koedinger.
\newblock Beyond accuracy: Embracing meaningful parameters in educational data mining.
\newblock In {\em Proceedings of the 17th International Conference on Educational Data Mining}, pages 203--210, 2024.

\bibitem{reckase200618_mirt}
M.~D. Reckase.
\newblock Multidimensional item response theory.
\newblock {\em Handbook of statistics}, 26:607--642, 2006.

\bibitem{ruan-2021-vdkt}
S.~Ruan, W.~Wei, and J.~Landay.
\newblock Variational deep knowledge tracing for language learning.
\newblock In {\em {LAK21}: 11th {International} {Learning} {Analytics} and {Knowledge} {Conference}}, {LAK21}, pages 323--332, New York, NY, USA, Apr. 2021. Association for Computing Machinery.

\bibitem{rupp-2010-diagnostics}
A.~A. Rupp, J.~Templin, and R.~A. Henson.
\newblock {\em Diagnostic measurement: {Theory}, methods, and applications}.
\newblock Diagnostic measurement: {Theory}, methods, and applications. The Guilford Press, New York, NY, US, 2010.

\bibitem{shen-2021-knowledgetracing}
S.~Shen, Q.~Liu, Z.~Huang, Y.~Zheng, M.~Yin, M.~Wang, and E.~Chen.
\newblock A survey of knowledge tracing: Models, variants, and applications.
\newblock {\em IEEE Transactions on Learning Technologies}, 17:1858–1879, 2021.

\bibitem{song-2023-cmncd}
L.~Song, M.~He, X.~Shang, C.~Yang, J.~Liu, M.~Yu, and Y.~Lu.
\newblock A deep cross-modal neural cognitive diagnosis framework for modeling student performance.
\newblock {\em Expert Syst. Appl.}, 230(C), Nov. 2023.

\bibitem{templin-2013-cdm}
J.~Templin and L.~Bradshaw.
\newblock Measuring the reliability of diagnostic classification model examinee estimates.
\newblock {\em Journal of Classification}, 30(2):251 -- 275, 2013.

\bibitem{verhelst-1993-dynamicRasch}
N.~D. Verhelst and C.~A.~W. Glas.
\newblock A dynamic generalization of the rasch model.
\newblock {\em Psychometrika}, 58(3):395--415, Sept. 1993.

\bibitem{walsh2018PPE1}
M.~M. Walsh, K.~A. Gluck, G.~Gunzelmann, T.~Jastrzembski, and M.~Krusmark.
\newblock Evaluating the theoretic adequacy and applied potential of computational models of the spacing effect.
\newblock {\em Cognitive science}, 42:644--691, 2018.

\bibitem{walsh2018PPE2}
M.~M. Walsh, K.~A. Gluck, G.~Gunzelmann, T.~Jastrzembski, M.~Krusmark, J.~I. Myung, M.~A. Pitt, and R.~Zhou.
\newblock Mechanisms underlying the spacing effect in learning: A comparison of three computational models.
\newblock {\em Journal of Experimental Psychology: General}, 147(9):1325, 2018.

\bibitem{wang-2021-hawkeskt}
C.~Wang, W.~Ma, M.~Zhang, C.~Lv, F.~Wan, H.~Lin, T.~Tang, Y.~Liu, and S.~Ma.
\newblock Temporal cross-effects in knowledge tracing.
\newblock In {\em Proceedings of the 14th ACM International Conference on Web Search and Data Mining}, WSDM '21, page 517–525, New York, NY, USA, 2021. Association for Computing Machinery.

\bibitem{wang-2019-longIRT}
C.~Wang and S.~W. Nydick.
\newblock On longitudinal item response theory models: {A} didactic.
\newblock {\em Journal of Educational and Behavioral Statistics}, 45(3):339--368, June 2019.
\newblock Publisher: American Educational Research Association.

\bibitem{wang-2023-dcd}
F.~Wang, Z.~Huang, Q.~Liu, E.~Chen, Y.~Yin, J.~Ma, and S.~Wang.
\newblock Dynamic cognitive diagnosis: An educational priors-enhanced deep knowledge tracing perspective.
\newblock {\em IEEE Trans. Learn. Technol.}, 16(3):306–323, June 2023.

\bibitem{wang-2019-neuralcd}
F.~Wang, Q.~Liu, E.~Chen, Z.~Huang, Y.~Chen, Y.~Yin, Z.~Huang, and S.~Wang.
\newblock Neural cognitive diagnosis for intelligent education systems, 2019.
\newblock arXiv:1908.08733.

\bibitem{wang-2018-dynamicCDM}
S.~Wang, Y.~Yang, S.~A. Culpepper, and J.~A. Douglas.
\newblock Tracking skill acquisition with cognitive diagnosis models: {A} higher-order, hidden markov model with covariates.
\newblock {\em Journal of Educational and Behavioral Statistics}, 43(1):57--87, Feb. 2018.
\newblock Publisher: American Educational Research Association.

\bibitem{wang-2013-dynamicIRT}
X.~Wang, J.~O. Berger, and D.~S. Burdick.
\newblock Bayesian analysis of dynamic item response models in educational testing.
\newblock {\em The Annals of Applied Statistics}, 7(1), Mar. 2013.

\bibitem{wang-2022-wrongDKT}
X.~Wang, Z.~Zheng, J.~Zhu, and W.~Yu.
\newblock What is wrong with deep knowledge tracing? attention-based knowledge tracing.
\newblock {\em Applied Intelligence}, 53(3):2850--2861, 2022.

\bibitem{wang-2020-varfa}
Z.~Wang, Y.~Gu, A.~Lan, and R.~Baraniuk.
\newblock {VarFA}: A variational factor analysis framework for efficient bayesian learning analytics, 2020.
\newblock arXiv:2005.13107.

\bibitem{Wilson2004}
M.~Wilson and P.~De~Boeck.
\newblock {\em Descriptive and explanatory item response models}, pages 43--74.
\newblock Springer New York, New York, NY, 2004.

\bibitem{wu-2020-vibo}
M.~Wu, R.~L. Davis, B.~W. Domingue, C.~Piech, and N.~Goodman.
\newblock Variational item response theory: Fast, accurate, and expressive, 2020.
\newblock arXiv:2002.00276.

\bibitem{yeung-2019-deepIRT}
C.-K. Yeung.
\newblock {Deep-IRT}: Make deep learning based knowledge tracing explainable using item response theory, 2019.
\newblock arXiv:1904.11738.

\bibitem{yeung-2018-dktPlus}
C.-K. Yeung and D.-Y. Yeung.
\newblock Addressing two problems in deep knowledge tracing via prediction-consistent regularization, 2018.
\newblock arXiv:1806.02180.

\bibitem{yin-2023-dtransformer}
Y.~Yin, L.~Dai, Z.~Huang, S.~Shen, F.~Wang, Q.~Liu, E.~Chen, and X.~Li.
\newblock Tracing knowledge instead of patterns: Stable knowledge tracing with diagnostic transformer.
\newblock In {\em Proceedings of the ACM Web Conference 2023}, WWW '23, page 855–864, New York, NY, USA, 2023. Association for Computing Machinery.

\bibitem{zhang-2016-dkvmn}
J.~Zhang, X.~Shi, I.~King, and D.-Y. Yeung.
\newblock Dynamic key-value memory networks for knowledge tracing, 2016.
\newblock arXiv:1611.08108.

\bibitem{zhang-2022-modifying}
Q.~Zhang, Z.~Chen, N.~Lalwani, and C.~MacLellan.
\newblock Modifying deep knowledge tracing for multi-step problems.
\newblock In {\em Proceedings of the 15th International Conference on Educational Data Mining}, 2022.

\bibitem{zhou-2025-dkt2}
Y.~Zhou, W.~Han, and J.~Chen.
\newblock {DKT2}: Revisiting applicable and comprehensive knowledge tracing in large-scale data, 2025.
\newblock arXiv:2501.14256.

\end{thebibliography}

% \section{Appendix}

% \balancecolumns

\end{document}